\documentclass{article}
\usepackage{ijcai24}

\usepackage{times}
\usepackage{soul}
\usepackage{url}
\usepackage[hidelinks]{hyperref}
\usepackage[utf8]{inputenc}
\usepackage[small]{caption}
\usepackage{graphicx}
\usepackage{amssymb}
\usepackage{geometry}
\usepackage{colortbl}
\usepackage{amsthm}
\usepackage{amsmath}
\usepackage{natbib}
\usepackage{booktabs}
\usepackage{algorithm}
\usepackage{algorithmic}
\usepackage[switch]{lineno}

\title{Extending Information Bottleneck Attribution to Video Sequences}

\author{
Veronika Solopova$^1$
\and
Lucas Schmidt$^1$\and
Dorothea Kolossa$^1$
\affiliations
$^1$Technische Universität Berlin\\
\emails
\{veronika.solopova,dorothea.kolossa\}@tu-berlin.de
}

\begin{document}

\maketitle

\begin{abstract}
    We introduce VIBA, a novel approach for explainable video classification by adapting Information Bottlenecks for Attribution (IBA) to video sequences. While most traditional explainability methods are designed for image models, our IBA framework addresses the need for explainability in temporal models used for video analysis. To demonstrate its effectiveness, we apply VIBA to video deepfake detection, testing it on two architectures: the Xception model for spatial features and a VGG11-based model for capturing motion dynamics through optical flow. Using a custom dataset that reflects recent deepfake generation techniques, we adapt IBA to create relevance and optical flow maps,  visually highlighting manipulated regions and motion inconsistencies. Our results show that VIBA generates temporally and spatially consistent explanations, which align closely with human annotations, thus providing interpretability for video classification and particularly for deepfake detection.
\end{abstract}

\section{Introduction}
The rising need for explainability in video classification models has prompted research into extending interpretability methods traditionally designed for image-based models to the temporal and spatial complexities of video sequences. Existing approaches in explainable AI (XAI) often focus on static image analysis, leaving a gap in interpretability for models that incorporate dynamic, time-dependent information critical in video applications. In this study, we adapt the Information Bottleneck for Attribution (IBA)~\citep{schulz2020restrictingflowinformationbottlenecks} method to video sequences, offering a new VIBA (Video Information Bottleneck Attribution) approach to produce visual explanations that address both spatial and temporal dimensions.
To illustrate our method, we apply VIBA in a deepfake detection task. Here, a significant challenge arises from the need to catch both subtle spatial manipulations and temporal inconsistencies, since detecting deepfakes with the human eye is becoming increasingly difficult as the quality of fake media improves. We test VIBA with two model architectures: Xception for capturing spatial features and a VGG11-based optical flow model for temporal motion dynamics. Using a dataset that incorporates recent deepfake generation techniques, we demonstrate the ability of VIBA to generate relevance and optical flow maps, effectively highlighting keyframes and motion patterns relevant to model predictions. We then analyse the effect of IBA on the model performance and the consistency of the highlighted regions over the length of the video. We also compare the IBA explanations to human annotations we collected from lay and expert annotators. Our code is available in an anonymous GitHub repository \footnote{\url{https://github.com/anonrep/IBA-for-Video-Sequences}}. The trained models will be made available upon acceptance.
This contribution provides a pathway to interpretable video models, supporting applications beyond deepfake detection and enhancing the transparency of temporal model decision-making.

\section{Related Work}

In recent years, deep learning models have significantly improved tasks like object detection and medical imaging, yet their ``black-box" nature complicates interpretability, especially in sensitive areas like healthcare \citep{MONTAVON2017211}. Explainable AI (XAI) seeks to address this by enhancing transparency in model decision-making.

 Relevance maps are central to explainable image processing, highlighting the areas most influential to model predictions. Grad-CAM~\citep{Selvaraju_2019}, a widely used technique, generates heatmaps based on gradient information from a model's final layers, helping identify key features in classifier decisions. Another popular method, Layer-wise Relevance Propagation (LRP)~\citep{binder2016layerwiserelevancepropagationneural}, back-propagates relevance scores to assign pixel importance, clarifying the contribution of different layers \citep{LapuschkinS}.
Integrated Gradients~\citep{sundararajan2017axiomaticattributiondeepnetworks} computes attributions by integrating gradients from a baseline to the actual input, offering precise relevance insights. Perturbation-based techniques, like Occlusion Sensitivity \citep{zeiler2013visualizingunderstandingconvolutionalnetworks}, evaluate importance by masking input regions, making them applicable as black-box methods.
Information Bottlenecks for Attribution (IBA)~\citep{schulz2020restrictingflowinformationbottlenecks}, which we used as a basis for this study, uses information bottlenecks, selectively limiting input information to assess regional importance, providing more robust explanations. 

XAI methods for video sequences are, however, a less explored field. 
\cite{Chen2021WhereAW} propose a space-time attention network for multi-modal settings that uncovers the synergistic dynamics of audio and visual data over both space and time, by localizing where the
relevant visual cues appear, and when the predicted sounds occur in videos. \cite{lee2024gradientbasedtimeseriesexplanationsspatiotemporal} use a transformer-based, spatiotemporal attention network (STAN) for gradient-based time-series explanations for video classification, producing salient frames of time-series data. In \cite{10.1145/3626772.3657664}, the authors adapt Grad-CAM to generate heatmaps highlighting regions of the video that significantly influence the model's predictions. 
 
 An application attracting a lot of attention for explainable models has been deepfake detection. \cite{9092227} adapted model-agnostic LRP and LIME approaches for this task. Using an adversarial attacks framework, \cite{Tsigos2024TowardsQE} compared the performance of Grad-CAM++ \citep{Chattopadhay_2018}, SHAP \citep{10.5555/3295222.3295230}, LIME \citep{10.1145/2939672.2939778}, and SOBOL \citep{hart2016efficientcomputationsobolindices}, where for this task LIME showed the best performance. \cite{GOWRISANKAR2024103684} evaluate XAI for deepfakes using an adversarial attack approach.
To the best of our knowledge, the benefits of IBA explanations for videos have not yet been investigated. However, multiple studies have shown that it provides a more stable estimation for images and text than similar attribution methods \citep{schulz2020restrictingflowinformationbottlenecks, Demir2021InformationBA, jiang-etal-2020-inserting}.
IBA's key advantage is its ability to quantify relevance in bits, providing a clear, interpretable measure for attribution, something gradient-based methods can struggle with due to numerical instability and sensitivity to model architecture \citep{adebayo2020sanitycheckssaliencymaps}. Additionally, IBA is a post-hoc method, applicable to any pre-trained black-box model without needing training data or internal parameters, while IBA’s iterative process enhances its robustness to over-attribution and ensures that only the most important features are emphasized, while also slightly improving the generalization of the explained model \citep{samek2015evaluatingvisualizationdeepneural}.  
\raggedbottom
\section{Methods}
In the following section, we describe the processing pipeline, detail the construction of the dataset used and the trainings we carried out, and explain the proposed adaptations to IBA for motion analysis.
\subsection{Model Architecture and Training}
    \begin{figure}[!h]
        \centering
        \includegraphics[scale=0.30]{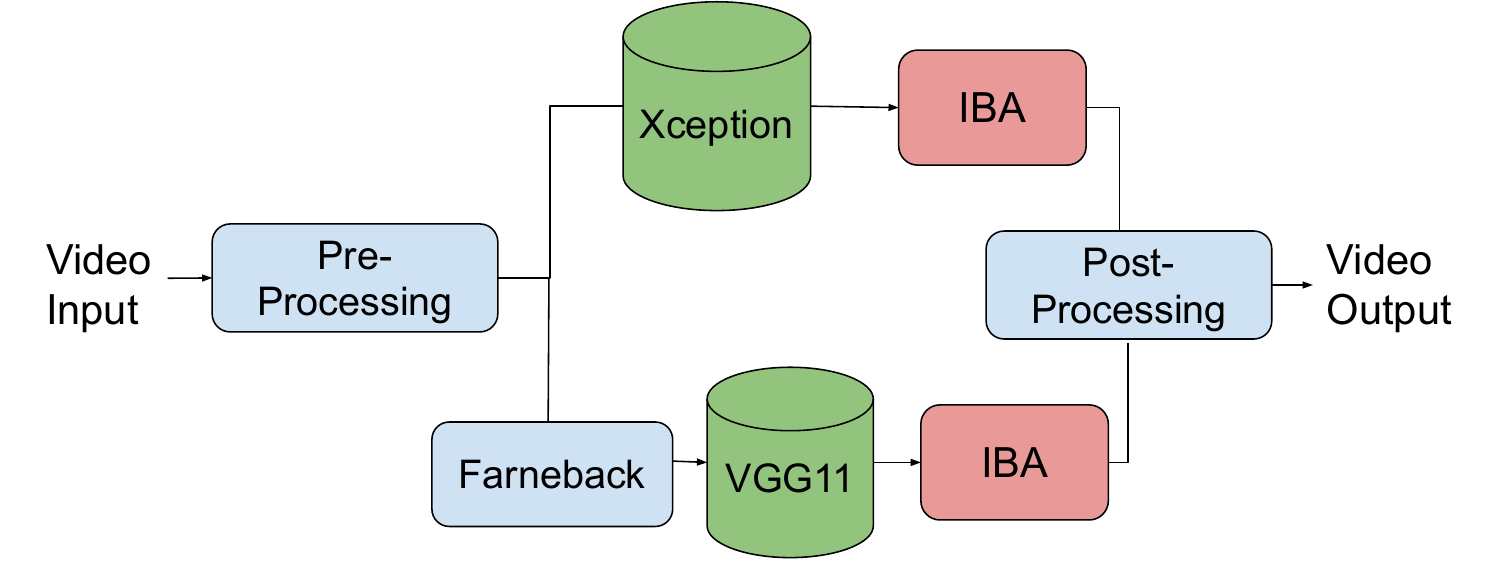}
        \caption{Implemented VIBA pipeline.}
        \label{fig:pipeline}
    \end{figure}

    We structured the process into two main analysis paths, as depicted in Figure~\ref{fig:pipeline}: the spatial analysis relies on a pre-trained Xception model, while our newly proposed temporal analysis is made possible by training a motion-artefact detection model using a VGG11-based optical flow model. This setup allows us to test IBA for both types of detection, comparing how well it explains features across different aspects of the
    videos.\\
      The Xception model architecture was initialized with publicly available pre-trained weights from \cite{liu2023deepfake}. 
    The artefact detection was trained using the VGG11 architecture on optical flow maps derived from manipulated and authentic video frames. To avoid overfitting, an early stopping mechanism was implemented based on the validation loss. The training process involved using over 10,000 pairs of frames, consisting of randomly selected keyframes and their subsequent frames from both real and deepfake videos. These pairs of frames were then converted into ca.~5000 optical flow maps.\\
    \subsection{Dataset Construction}
    For our use case implementation, we created a deepfake detection dataset\footnote{Due to licensing constraints, the dataset can not be made publicly available, but it can be shared with vetted academics for reproducibility purposes upon request.} combining many types of manipulated and authentic videos from established sources such as FaceForensics++ \citep{FF++/Xception}, Celeb-DF \citep{CDF}, Deepfake Detection Challenge (DFDC) \citep{DFDC}, Deepfake Detection Dataset (DFD) \citep{DFD}, DeeperForensics \citep{DFo}, FakeAVCeleb \citep{fakeAVCeleb}, AV-Deepfake1M \citep{1M} and the Korean Deepfake Detection Dataset (KoDF) \citep{KoDF}. Approximately 50 videos were sampled from each dataset, ensuring diversity across manipulation methods. The dataset was refined with 30 state-of-the-art deepfakes from flagged and attested fake YouTube videos and short clips from the TV show ``Deep Fake Neighbour Wars". The dataset was randomly split into three parts: 70\% was used to train the VGG11-based model (\emph{Train} dataset), and 15\% for its validation. The remaining 15\% of the dataset was used to test both deepfake detection models and to evaluate the VIBA explanations (further referred to as the \emph{Evaluation} dataset).
 \begin{table}
    \centering
    \caption{Distribution of selected videos from various datasets with different deepfake techniques.}
    \renewcommand{\arraystretch}{1.1} 
    \setlength{\tabcolsep}{6pt} 
    \begin{tabular}{l c}
        \toprule
        \textbf{Dataset} & \textbf{\# Videos} \\
        \midrule
        FaceForensics++ & 50 \\
        Celeb-DF        & 56 \\
        DFDC            & 50 \\
        DFD             & 50 \\
        FakeAVCeleb     & 46 \\
        AV-Deepfake1M   & 54 \\
        KoDF            & 48 \\
        YouTube Videos  & 18 \\
        TV show ``Deep Fake Neighbour Wars" & 6 \\
        \bottomrule
    \end{tabular}
    \label{tab:dataset_videos}
\end{table}

    Preprocessing followed a pipeline to standardize input data for both models. Videos were first split into individual frames, and keyframe extraction following ~\cite{libaceta2023keyframe} was applied to reduce redundancy and computational load, ensuring that only the most informative frames were retained. Faces were then detected and cropped from each frame using the Dlib face recognition method \citep{10.5555/1577069.1755843} to focus on the primary region of manipulation and remove extraneous background elements. The cropped faces were resized to meet the input dimensions of each model: 299×299 pixels for Xception and 256×256 pixels for VGG11. For the VGG11-based model, dense optical flow maps were generated using the Farneback algorithm \citep{Gunnar_Farnebäck}. These maps captured pixel motion between consecutive frames, providing a robust representation of temporal dynamics.\\
    \subsection{IBA for Motion Analysis}
  IBA applies the Information Bottleneck principle from \cite{tishby2000informationbottleneckmethod} to enhance explainability in neural networks \citep{schulz2020restrictingflowinformationbottlenecks}. The main concept is to insert a bottleneck into the intermediate layers of the network and iteratively reduce the information flow through these layers. This is done by adding noise to feature maps, limiting the effect of certain input regions, and observing how the model's output is altered. By assessing the sensitivity of the output to various input regions, IBA identifies the areas that have the greatest influence on the final decision in the following way:
  
            In a typical deep neural network, let \( R \) represent the feature map output of a given layer, with dimensions \((N, C, H, W)\), where \( N \) is catch size, \( C \)is number of channels (or features), and  \( H \) and \( W \) are height and width of the feature map. Let \( \epsilon \sim \mathcal{N}(\mu_R, \sigma_R^2) \) be Gaussian noise with the same dimensions as \( R \) added to this feature map. \( \mu_R \) and \( \sigma_R \) are mean and standard deviation of \( R \), calculated over specific axes. 
            The modified representation \( Z \) is calculated as a linear combination of the original feature map \( R \) and the noise \( \epsilon \) with the combination weighted by an attention parameter \( \lambda(X) \):
            
            \begin{equation}
            Z = \lambda(X) R + (1 - \lambda(X)) \epsilon,
            \end{equation}
            
            where X is the input data, \( \lambda(X) \)  has the same dimensionality as \( R \) and is applied element-wise, controlling how much information from \( R \) is retained and \( (1 - \lambda(X)) \) represents the amount of noise introduced. 

            The objective is to find the optimal \( \lambda(X) \) that best retains the model's prediction capability while keeping the information flow minimal.
\begin{equation}
\min_{\lambda(X)} \mathcal{L}(\theta, \lambda(X)) + \beta I(R; Z),
\end{equation}

where \( \mathcal{L}(\theta, \lambda(X)) \) is the model's loss function (e.g., cross-entropy loss); \( I(R; Z) \) is the mutual information between \( R \) and \( Z \), which quantifies how much information about \( R \) is retained in \( Z \). Smaller \( I(R; Z) \) implies more noise is added and \( \beta \) is a trade-off parameter controlling the balance between prediction capability and information flow.
    
    \subsubsection{Injection Layer Identification}
    We identified the appropriate layer for bottleneck injection based on a balance between feature abstraction and spatial detail, guided by the methodology in \cite{schulz2020restrictingflowinformationbottlenecks}.
    The selection process was primarily qualitative, focusing on the clarity and interpretability of relevance maps generated at different stages of the model's architecture. We systematically tested several injection points within the Xception model, including block 4, bn3, conv3, block 12, and bn4 (see Table \ref{tab:Xception} and Table~\ref{tab:block} in Appendix \ref{app:IBA_Xception} for layer definitions and sizes of the Xception architecture), each representing progressively deeper layers, where the feature maps evolve from detailed spatial patterns to highly abstract representations.
    
    This assessment involved generating relevance maps for the same set of real and deepfake images across these layers and visually comparing their clarity and ability to highlight subtle manipulation artefacts. Consistently, block 4 produced the most informative and human-interpretable heatmaps, as it retained fine-grained spatial details such as edges, textures, and local patterns critical for deepfake detection. Deeper layers like bn3 and block 12 continued to capture relevant information, but showed increasing abstraction of the feature maps, making the heatmaps less spatially detailed but still useful for highlighting broader patterns. Conversely, layers such as conv3 and bn4, closer to the network's final decision layers, produced highly abstract and diffused relevance maps, losing essential spatial information and becoming less informative for localized deepfake detection. You can find sample frame relevance maps from each injection block in Appendix \ref{xception:choice}, Figures \ref{fig:block4} to \ref{fig:bn4}.
    
    In repeating these experiments with the VGG11 model, we saw similar outcomes, where earlier layers (Layer 9 and 12) also retained finer temporal details, while deeper layers, like Layer 16, became too abstract for effective visual analysis (see sample optical flow maps from each layer in Appendix \ref{vgg:choice}, Figures \ref{fig:Batch Normalization2} to \ref{fig:Batch Normalization4}). These observations reinforce the general pattern that earlier convolutional layers, where spatial and temporal details remain rich, strike a better balance between preserving interpretability and capturing essential patterns for manipulation detection.\\
    Therefore, based on these qualitative comparisons, we selected block 4 in the Xception model and layer 9 for VGG11 as the optimal bottleneck injection points for all subsequent experiments.
    \\
    \subsubsection{Bottleneck type selection}
    A key decision for IBA-based attribution is whether to use a per-sample bottleneck or a readout bottleneck. The per-sample bottleneck fine-tunes noise injection for individual inputs, generating relevance maps tailored to specific examples, while the readout bottleneck uses a pre-trained neural network to predict noise for new inputs and is more scalable. We tested both methods using the Xception model, since both architectures share a hierarchical convolutional design, aligning with the principles of feature abstraction in CNNs, and both models also exhibited a similar pattern of optimal bottleneck placement in earlier layers.
    
    After training with \textit{Train}
dataset for 10 epochs, the relevance maps generated by the readout bottleneck lacked consistency and clarity in comparison to those produced by the per-sample bottleneck. This also aligns with the original recommendations of \cite{schulz2020restrictingflowinformationbottlenecks}. The generated relevance maps were reassembled into video sequences, overlaying heatmaps on the original frames to provide a dynamic visualization of regions relevant to decision making. These visualizations demonstrate how the models detected discrepancies across keyframes, offering a view of spatial and temporal anomalies. 
     \\
     \subsection{Evaluation}
     We evaluate the consistency of our explanations using three tests:
     \begin{enumerate}
         \item \textbf{Comparative baseline testing.}
         We verify the performance of the models on the task of our choice, with and without IBA injection on the test set, to see if the injected noise has a negative influence on the predictive accuracy. We use Accuracy, Precision, Recall and Expected Calibration Error (ECE) \citep{guo2017calibration}. 
       \item  \textbf{Saliency map consistency.} To analyze the consistency of the saliency maps we produce, we chose
        three metrics: Intersection over Union (IoU), Temporal Consistency Score (TCS), and Region Persistence Index (RPI). We compute the mean of these metrics over the generated VIBA explanations for the test set.
         IoU measures the spatial overlap between two binary masks, showing how similar the highlighted regions are between consecutive frames. Scores closer to 1 indicate greater consistency. 

         TCS measures the proportion of frames where regions stay consistently highlighted over time. Scores closer to 1 indicate higher consistency throughout the entire video sequence.\\
         \begin{equation}
    TCS = \frac{1}{N} \sum_{i=1}^{N} \frac{\sum_{x,y} M_i(x, y)}{W \times H}
\end{equation}
where $N$ is the total number of frames; $M_i(x, y)$ is the binary mask value for pixel $(x, y)$ in frame $i$; $W$ and $H$ are the width and height of the image.\\
$RPI$ measures the average movement (in pixels) of the centroid of the highlighted region across frames. 
\begin{equation}
    RPI = \frac{\frac{1}{N-1} \sum_{i=2}^{N} \| C_i - C_{i-1} \|}{\sqrt{W^2 + H^2}}
\end{equation}
where $N$ is the number of frames; $C_i$ is the centroid of the binary mask for frame $i$; $\| C_i - C_{i-1} \|$ is the Euclidean distance between the centroids of consecutive frames; 
As we have a stable size of frames (924, 924), we calculate the upper limit of the centroid movement and normalize the values from 0 to 1, with lower values indicating more consistency, and 0 - perfect stability.\\
\item \textbf{Ablation testing with known artefacts.}
To see whether VIBA and the models are picking up on essential cues or just spurious correlations, we asked 8 annotators (two senior researchers with PhDs in Engineering, five PhD students in Computer Science, and a trained journalist to annotate 27 deepfake videos of the Evaluation dataset with regions they found most indicative when identifying a deepfake. We used the top three most frequent answers to calculate the F1-score (macro), Precision, Recall and Overlap (Szymkiewicz–Simpson Coefficient). Additionally, we measured the percentage of cases where the most commonly selected region by human annotators (majority vote) was present in the VIBA explanation. We did not offer the annotators any real images, as they do not contain artefacts to annotate.
     \end{enumerate}
\raggedbottom

\section{Results}
    
Our results, presented in Table \ref{tab:model_performance}, show that VIBA barely alters the overall accuracy, but improves the quality of attributions. We also observe a slight improvement in the ECE of the Xception model, where VIBA might reduce overconfidence by ensuring the model's predictions are based only on the most relevant information.
The Xception model shows slightly better performance than the VGG model across metrics, but overall results are comparable, which allows us to optimally evaluate VIBA explanations for two different architectures.
\begin{table}
    \centering
    \caption{Values in \%. Performance of Xception and VGG11 on deepfake detection. A plus represents VIBA injection and a minus means standard model without VIBA.}
    \renewcommand{\arraystretch}{1.1} 
    \setlength{\tabcolsep}{4pt} 
    \begin{tabular}{lcccc}
        \toprule
        \textbf{Model} & \textbf{Acc.} & \textbf{Precision} & \textbf{Recall} & \textbf{ECE} \\
        \midrule
        Xception -  & 81.51 & 77.49 & 89.34 & 0.1216 \\
        Xception +  & 81.47 & 77.43 & 89.34 & 0.1210 \\
        VGG -       & 79.96 & 75.42 & 88.95 & 0.1244 \\
        VGG +       & 79.87 & 75.31 & 88.95 & 0.1246 \\
        \bottomrule
    \end{tabular}
    \label{tab:model_performance}
\end{table}
\begin{table}[!ht]
    \centering
    \caption{Mean evaluation metrics of saliency map consistency for VIBA explanations of VGG \& Xception models. We show separate results for deepfakes and real videos.}
    \renewcommand{\arraystretch}{1.1} 
    \setlength{\tabcolsep}{4pt} 
    \begin{tabular}{lccc}
        \toprule
        \textbf{Model} & \textbf{IoU} & \textbf{TCS} & \textbf{RPI} \\
        \midrule
        Xception Fake & 0.8226 & 0.4306 & 0.0255\\
        Xception Real & 0.8320 & 0.4086 & 0.0250 \\
        VGG Fake      & 0.7633 & 0.8611 & 0.0341 \\
        VGG Real      & 0.8001 & 0.8766 & 0.0267 \\
        \bottomrule
    \end{tabular}
    \label{tab:saliency_consistency}
\end{table}\raggedbottom
Overall, saliency maps of VIBA explanations show slightly better consistency for real examples (see Table \ref{tab:saliency_consistency}). While highlighted regions of Xception are on average more consistent over consecutive frames, the temporal consistency score of VGG11 with its optical flow maps is more than twice higher over the whole video sequence. At the same time, it is natural for different regions of deepfake manipulations to become more pronounced throughout the course of the video, and this may indicate that Xception is more sensitive to such changes.\\
In terms of the centroid movement of the highlighted regions, the explanation generated for both models seems rather stable with numbers close to 0, with slightly less consistency observable for deepfakes in the case of the VGG-based model. \\
As we can observe in Table \ref{tab:f1_comparison}, both models show moderate agreement with human annotations, overlapping with human labels in more than 50\% of videos. The region most important by the majority vote of human annotations was also identified by the model in 40\% of the cases for the VGG model and around 63\% for Xception. Based on the higher recall of the models compared to precision, they cover a considerable amount of the regions humans marked as indicative but also highlight other regions.  For instance, the most frequently identified areas by human annotators were lips and mouth area as well as brows, eyes and forehead, which is also true for explanations produced for Xception outputs.\\
For VGG11 the most frequently highlighted area is the eyes, brows and forehead, which is also the second most important for Xception.
\begin{figure}
    \centering
    \includegraphics[width=\columnwidth]{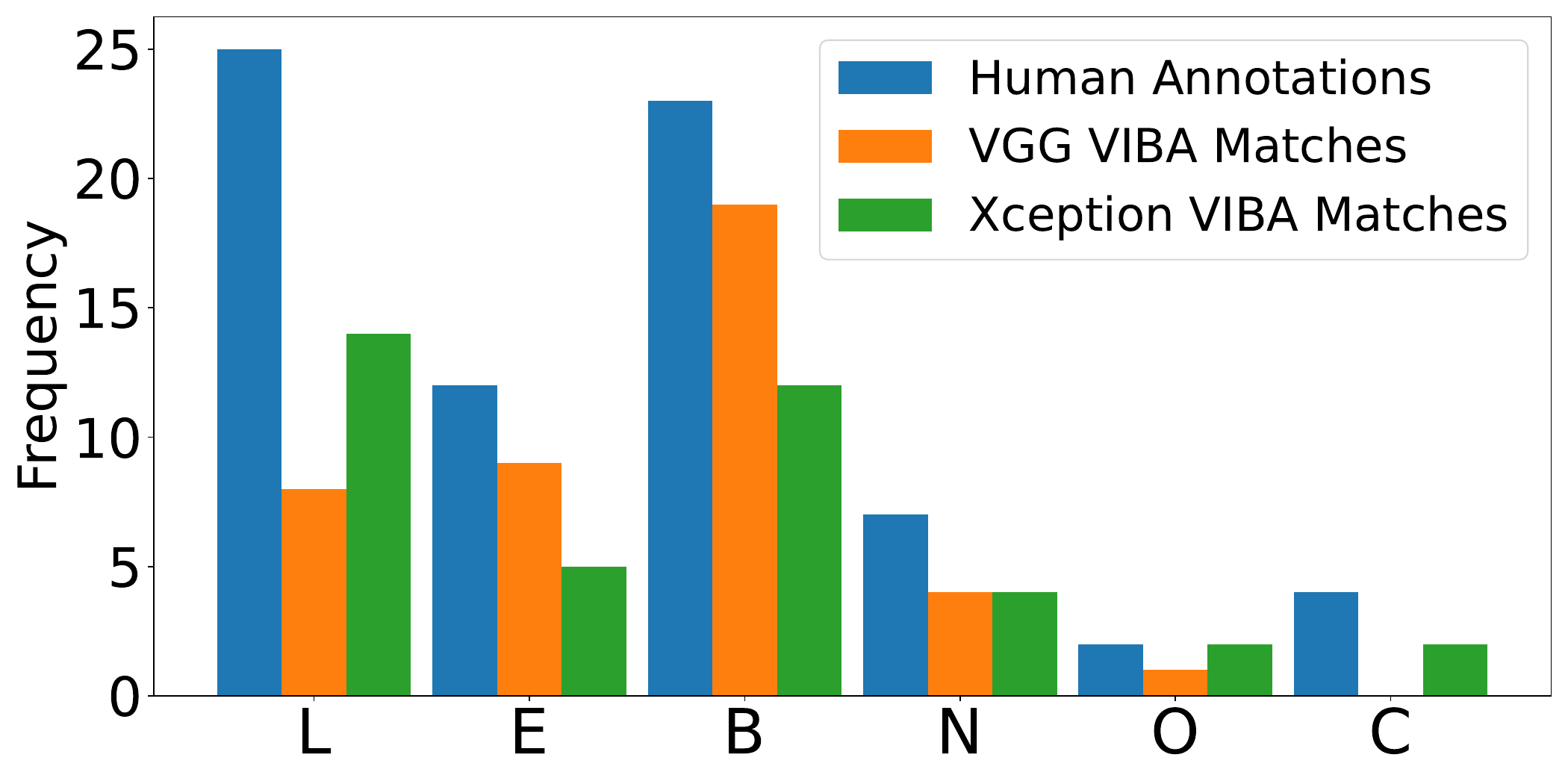}
    \caption{Comparison of frequencies for human annotations, VGG and Xception VIBA matches for different face regions. Abbreviations: L: Lips \& Mouth, E: Deepfake Edges, B: Brows, Eyes \& Forehead, N: Nose, Cheeks \& Ears, O: Outside of Face, C: Chin \& Neck.}

    \label{fig:ablation_study}
\end{figure}
\begin{table}
    \centering
    \caption{Values in \%: F1 Score, Precision, Recall, Overlap coefficient \& Top 1 match from human majority vote between VIBA-highlighted regions of deepfakes and human annotations for Xception \& VGG model predictions.}
    \renewcommand{\arraystretch}{1.1} 
    \setlength{\tabcolsep}{4pt} 
    \begin{tabular}{l c c c c c}
        \toprule
        \textbf{Model} & \textbf{F1}   & \textbf{Prec.}& \textbf{Recall}& \textbf{Overlap}&\textbf{Top 1}\\
        \midrule
        Xception   & 45.42 & 40.76&51.28&51.92&62.96\\
        VGG        & 52.72 & 50.00&55.76&57.69&40.74\\
        \bottomrule
    \end{tabular}
    \label{tab:f1_comparison}
\end{table}
IBA explanations for VGG matched well the brows, eyes and forehead area, and visible deepfake edges. It did not attribute any importance to the chin and neck area, even when human annotators did.
In terms of explanations for the Xception model, lips and mouth were identified well and better than in the case of the VGG model, while eyes, brows and forehead were also identified moderately well. It was also slightly better at identifying when regions outside of the face were suggestive of manipulations. In contrast to the VGG model, Xception explanations often focused on the chin and neck, but this often does not coincide with human annotations. Explanations for both models similarly often identified nose, ears and cheek areas, but less frequently than human annotators.
In terms of qualitative findings within the defined face parts, Xception often focused on the nose and forehead area, and almost never on the eyes. Eyes and ears are highly frequent for the VGG model. Explanations for real videos and deepfakes focus on similar areas. For fake videos, the Xception model tends to focus on the forehead when analyzing the ``eyes, brows, and forehead" zone. In contrast, for real videos, the explanations provided by the VIBA highlight the eyes more precisely. Similarly, in the ``nose, cheeks, and ears" zone, the model emphasizes the nose and ears for fake videos. For real videos, however, it focuses more on the cheeks. Interestingly, in the case of the VGG model, which never focuses on the nose and chin when it comes to fake videos, VIBA does highlight these regions in real videos.

\section{Discussion}

In this study, we demonstrated how information bottleneck attribution can be adapted to video sequence explanations. We showed the application of this proposed VIBA approach to the timely task of deepfake detection by providing explanations fitted to two substantially different architectures. 
Neither of the models that we implemented to illustrate VIBA---the standard Xception model and an optical-flow-based VGG11 architecture---showed a statistically significant drop in performance. The near-identical recall and precision values suggest the implementation is stable and does not introduce randomness or performance degradation. The ECE even improved slightly with VIBA, indicating the model is better calibrated with more reliable probability outputs.
Our analysis shows that VIBA explanations produced for both models show high temporal and spatial consistency of the saliency maps. The regions important for the human annotators are moderately well picked up by the model, while indicative features outside of the face are not prevalent, indicating that features that are important for the model but not for human annotators, are still found in the face and might still pick up artefacts the human eye misses. In our qualitative analysis, we also observed notable differences in the prevalence of important regions between real videos and deepfakes, indicating the presence of class-specific features.
The optical flow maps for the VGG11-based model capture the motion patterns between frames. This approach is effective in identifying subtle temporal inconsistencies in fake videos. However, as these relevance maps focus more on motion dynamics, they are less intuitive for human interpretation than the static frame-based relevance maps obtained for the Xception model. 
Comparing VIBA to popular Grad-CAM explanations, Grad-CAM is limited to the final convolutional layer, which can lead to coarse explanations, as higher layers in CNNs often leverage more abstract features, while overreliance on backpropagated gradients leads to incomplete or misleading visual explanations, especially in high confidence predictions \citep{schulz2020restrictingflowinformationbottlenecks}. IBA directly controls the amount of information retained in the explanation by introducing a bottleneck, ensuring that only the most relevant features for the decision are highlighted, while not being tied to CNN architectures, and thus applicable to a broader range of deep learning models. Its mechanism of perturbing activations and focusing on how information flows through the network rather than relying purely on gradients makes it additionally generalizable, producing more fine-grained heatmaps, and reducing the risk of gradient saturation. 
 Despite its strengths, IBA has certain limitations, namely its computational complexity, and its reliance on noise to restrict information flow. While this approach provides a strong mechanism for attribution, it may not always align with human intuition of relevance, as we can see by only moderate agreement with human annotators in our case. For instance,  regions containing detailed information that are not directly related to the prediction may be down-weighted by IBA, even though humans might perceive these details as important for understanding the overall frame \citep{hendricks2018groundingvisualexplanations}. XAI methods are particularly effective in human-in-the-loop (HIL) scenarios because they enable human experts to interpret model outputs, identify potential errors, and make informed decisions. This is especially relevant for deepfake detection, where nuanced judgments and context are often required to distinguish subtle manipulations.

\section{Conclusion}
We extended the Information Bottleneck for Attribution (IBA) method to explainable video classification, applying it to spatial (Xception) and temporal (VGG11 optical flow) models for deepfake detection. VIBA generated consistent and detailed relevance and optical flow maps, enhancing interpretability without affecting model performance, with results highlighting its versatility across architectures and potential for broader applications in video analysis tasks.
\section*{Ethical Statement}
 Our models should not be used as a stand-alone fact-checking solution, but they can be used as weak annotation tools, or as a help for human fact-checkers, with the final decision to be made by a human. 
 
\bibliographystyle{named}
\bibliography{ijcai24}

\begin{thebibliography}{}

\bibitem[\protect\citeauthoryear{Adebayo \bgroup \em et al.\egroup }{2020}]{adebayo2020sanitycheckssaliencymaps}
Julius Adebayo, Justin Gilmer, Michael Muelly, Ian Goodfellow, Moritz Hardt, and Been Kim.
\newblock Sanity checks for saliency maps, 2020.

\bibitem[\protect\citeauthoryear{Binder \bgroup \em et al.\egroup }{2016}]{binder2016layerwiserelevancepropagationneural}
Alexander Binder, Grégoire Montavon, Sebastian Bach, Klaus-Robert Müller, and Wojciech Samek.
\newblock Layer-wise relevance propagation for neural networks with local renormalization layers, 2016.

\bibitem[\protect\citeauthoryear{Cai \bgroup \em et al.\egroup }{2023}]{1M}
Zhixi Cai, Shreya Ghosh, Aman~Pankaj Adatia, Munawar Hayat, Abhinav Dhall, and Kalin Stefanov.
\newblock Av-deepfake1m: A large-scale {LLM}-driven audio-visual deepfake dataset, 2023.

\bibitem[\protect\citeauthoryear{Chattopadhay \bgroup \em et al.\egroup }{2018}]{Chattopadhay_2018}
Aditya Chattopadhay, Anirban Sarkar, Prantik Howlader, and Vineeth~N Balasubramanian.
\newblock Grad-cam++: Generalized gradient-based visual explanations for deep convolutional networks.
\newblock In {\em 2018 IEEE Winter Conference on Applications of Computer Vision (WACV)}. IEEE, March 2018.

\bibitem[\protect\citeauthoryear{Chen \bgroup \em et al.\egroup }{2021}]{Chen2021WhereAW}
Yanbei Chen, Thomas Hummel, A.~Sophia Koepke, and Zeynep Akata.
\newblock Where and when: Space-time attention for audio-visual explanations.
\newblock {\em ArXiv}, abs/2105.01517, 2021.

\bibitem[\protect\citeauthoryear{Demir \bgroup \em et al.\egroup }{2021}]{Demir2021InformationBA}
Ugur Demir, Ismail Irmakci, Elif Keles, Ahmet~Can Topçu, Ziyue Xu, Concetto Spampinato, Sachin~R. Jambawalikar, Evrim~B Turkbey, Baris~I Turkbey, and Ulas Bagci.
\newblock Information bottleneck attribution for visual explanations of diagnosis and prognosis.
\newblock {\em Machine learning in medical imaging. MLMI}, 12966:396--405, 2021.

\bibitem[\protect\citeauthoryear{Dolhansky \bgroup \em et al.\egroup }{2020}]{DFDC}
Brian Dolhansky, Joanna Bitton, Ben Pflaum, Jikuo Lu, Russ Howes, Menglin Wang, and Cristian~Canton Ferrer.
\newblock The deepfake detection challenge {(DFDC)} dataset, 2020.

\bibitem[\protect\citeauthoryear{Dufour and Gully}{}]{DFD}
Nick Dufour and Andrew Gully.
\newblock Google research.
\newblock \url{https://ai.googleblog.com/2019/09/contributing- data-to-deepfake-detection.html /}.
\newblock [Online; accessed 2024-04-24].

\bibitem[\protect\citeauthoryear{Farnebäck}{2003}]{Gunnar_Farnebäck}
Gunnar Farnebäck.
\newblock Two-frame motion estimation based on polynomial expansion.
\newblock In {\em Image Analysis, Proceedings of the Scandinavian Conference on Image Analysis (SCIA)}, volume 2749, pages 363--370, 06 2003.

\bibitem[\protect\citeauthoryear{Gowrisankar and Thing}{2024}]{GOWRISANKAR2024103684}
Balachandar Gowrisankar and Vrizlynn~L.L. Thing.
\newblock An adversarial attack approach for explainable {AI} evaluation on deepfake detection models.
\newblock {\em Computers \& Security}, 139:103684, 2024.

\bibitem[\protect\citeauthoryear{Guo \bgroup \em et al.\egroup }{2017}]{guo2017calibration}
Chuan Guo, Geoff Pleiss, Yu~Sun, and Kilian~Q Weinberger.
\newblock On calibration of modern neural networks.
\newblock In {\em Proceedings of the 34th International Conference on Machine Learning (ICML)}, pages 1321--1330. PMLR, 2017.

\bibitem[\protect\citeauthoryear{Hart \bgroup \em et al.\egroup }{2016}]{hart2016efficientcomputationsobolindices}
Joseph~L. Hart, Alen Alexanderian, and Pierre~A. Gremaud.
\newblock Efficient computation of {Sobol}' indices for stochastic models, 2016.

\bibitem[\protect\citeauthoryear{Hendricks \bgroup \em et al.\egroup }{2018}]{hendricks2018groundingvisualexplanations}
Lisa~Anne Hendricks, Ronghang Hu, Trevor Darrell, and Zeynep Akata.
\newblock Grounding visual explanations, 2018.

\bibitem[\protect\citeauthoryear{Ibaceta}{2023}]{libaceta2023keyframe}
Joel Ibaceta.
\newblock video-keyframe-detector, 2023.
\newblock Accessed: 2024-09-11.

\bibitem[\protect\citeauthoryear{Jiang \bgroup \em et al.\egroup }{2020a}]{DFo}
Liming Jiang, Ren Li, Wayne Wu, Chen Qian, and Chen~Change Loy.
\newblock Deeperforensics-1.0: A large-scale dataset for real-world face forgery detection, 2020.

\bibitem[\protect\citeauthoryear{Jiang \bgroup \em et al.\egroup }{2020b}]{jiang-etal-2020-inserting}
Zhiying Jiang, Raphael Tang, Ji~Xin, and Jimmy Lin.
\newblock {I}nserting {I}nformation {B}ottlenecks for {A}ttribution in {T}ransformers.
\newblock In Trevor Cohn, Yulan He, and Yang Liu, editors, {\em Findings of the Association for Computational Linguistics: EMNLP 2020}, pages 3850--3857, Online, November 2020. Association for Computational Linguistics.

\bibitem[\protect\citeauthoryear{Khalid \bgroup \em et al.\egroup }{2022}]{fakeAVCeleb}
Hasam Khalid, Shahroz Tariq, Minha Kim, and Simon~S. Woo.
\newblock {FakeAVCeleb}: A novel audio-video multimodal deepfake dataset, 2022.

\bibitem[\protect\citeauthoryear{King}{2009}]{10.5555/1577069.1755843}
Davis~E. King.
\newblock Dlib-ml: A machine learning toolkit.
\newblock {\em J. Mach. Learn. Res.}, 10:1755–1758, December 2009.

\bibitem[\protect\citeauthoryear{Kwon \bgroup \em et al.\egroup }{2021}]{KoDF}
Patrick Kwon, Jaeseong You, Gyuhyeon Nam, Sungwoo Park, and Gyeongsu Chae.
\newblock {KoDF}: A large-scale {Korean} deepfake detection dataset, 2021.

\bibitem[\protect\citeauthoryear{Lapuschkin \bgroup \em et al.\egroup }{2015}]{LapuschkinS}
Sebastian Lapuschkin, Alexander Binder, Grégoire Montavon, Frederick Klauschen, Klaus-Robert Müller, and Wojciech Samek.
\newblock On pixel-wise explanations for non-linear classifier decisions by layer-wise relevance propagation.
\newblock {\em PLoS ONE}, 10:e0130140, 07 2015.

\bibitem[\protect\citeauthoryear{Lee}{2024}]{lee2024gradientbasedtimeseriesexplanationsspatiotemporal}
Min~Hun Lee.
\newblock Towards gradient-based time-series explanations through a spatiotemporal attention network, 2024.

\bibitem[\protect\citeauthoryear{Li \bgroup \em et al.\egroup }{2019}]{CDF}
Yuezun Li, Xin Yang, Pu~Sun, Honggang Qi, and Siwei Lyu.
\newblock {Celeb-DF}: A new dataset for deepfake forensics, 09 2019.

\bibitem[\protect\citeauthoryear{Liu}{2023}]{liu2023deepfake}
Honggu Liu.
\newblock Deepfake detection.
\newblock \url{https://github.com/HongguLiu/Deepfake-Detection}, 2023.
\newblock Accessed: 2024-09-11.

\bibitem[\protect\citeauthoryear{Lundberg and Lee}{2017}]{10.5555/3295222.3295230}
Scott~M. Lundberg and Su-In Lee.
\newblock A unified approach to interpreting model predictions.
\newblock In {\em Proceedings of the 31st International Conference on Neural Information Processing Systems}, page 4765–4774, Red Hook, NY, USA, 2017. Curran Associates Inc.

\bibitem[\protect\citeauthoryear{Malolan \bgroup \em et al.\egroup }{2020}]{9092227}
Badhrinarayan Malolan, Ankit Parekh, and Faruk Kazi.
\newblock Explainable deep-fake detection using visual interpretability methods.
\newblock In {\em 2020 3rd International Conference on Information and Computer Technologies (ICICT)}, pages 289--293, 2020.

\bibitem[\protect\citeauthoryear{Montavon \bgroup \em et al.\egroup }{2017}]{MONTAVON2017211}
Grégoire Montavon, Sebastian Lapuschkin, Alexander Binder, Wojciech Samek, and Klaus-Robert Müller.
\newblock Explaining nonlinear classification decisions with deep {Taylor} decomposition.
\newblock {\em Pattern Recognition}, 65:211--222, 2017.

\bibitem[\protect\citeauthoryear{Ribeiro \bgroup \em et al.\egroup }{2016}]{10.1145/2939672.2939778}
Marco~Tulio Ribeiro, Sameer Singh, and Carlos Guestrin.
\newblock "{Why} should {I} trust you?": {Explaining} the predictions of any classifier.
\newblock In {\em Proceedings of the 22nd ACM SIGKDD International Conference on Knowledge Discovery and Data Mining}, KDD '16, page 1135–1144, New York, NY, USA, 2016. Association for Computing Machinery.

\bibitem[\protect\citeauthoryear{Rössler \bgroup \em et al.\egroup }{2019}]{FF++/Xception}
Andreas Rössler, Davide Cozzolino, Luisa Verdoliva, Christian Riess, Justus Thies, and Matthias Nießner.
\newblock Faceforensics++: Learning to detect manipulated facial images, 2019.

\bibitem[\protect\citeauthoryear{Samek \bgroup \em et al.\egroup }{2015}]{samek2015evaluatingvisualizationdeepneural}
Wojciech Samek, Alexander Binder, Grégoire Montavon, Sebastian Bach, and Klaus-Robert Müller.
\newblock Evaluating the visualization of what a deep neural network has learned, 2015.

\bibitem[\protect\citeauthoryear{Schulz \bgroup \em et al.\egroup }{2020}]{schulz2020restrictingflowinformationbottlenecks}
Karl Schulz, Leon Sixt, Federico Tombari, and Tim Landgraf.
\newblock Restricting the flow: Information bottlenecks for attribution, 2020.

\bibitem[\protect\citeauthoryear{Selvaraju \bgroup \em et al.\egroup }{2019}]{Selvaraju_2019}
Ramprasaath~R. Selvaraju, Michael Cogswell, Abhishek Das, Ramakrishna Vedantam, Devi Parikh, and Dhruv Batra.
\newblock Grad-cam: Visual explanations from deep networks via gradient-based localization.
\newblock {\em International Journal of Computer Vision}, 128(2):336–359, October 2019.

\bibitem[\protect\citeauthoryear{Sundararajan \bgroup \em et al.\egroup }{2017}]{sundararajan2017axiomaticattributiondeepnetworks}
Mukund Sundararajan, Ankur Taly, and Qiqi Yan.
\newblock Axiomatic attribution for deep networks, 2017.

\bibitem[\protect\citeauthoryear{Tishby \bgroup \em et al.\egroup }{2000}]{tishby2000informationbottleneckmethod}
Naftali Tishby, Fernando~C. Pereira, and William Bialek.
\newblock The information bottleneck method, 2000.

\bibitem[\protect\citeauthoryear{Tsigos \bgroup \em et al.\egroup }{2024}]{Tsigos2024TowardsQE}
Konstantinos Tsigos, Evlampios Apostolidis, Spyridon Baxevanakis, Symeon Papadopoulos, and Vasileios Mezaris.
\newblock Towards quantitative evaluation of explainable {AI} methods for deepfake detection.
\newblock {\em Proceedings of the 3rd ACM International Workshop on Multimedia AI against Disinformation}, 2024.

\bibitem[\protect\citeauthoryear{Vanneste \bgroup \em et al.\egroup }{2024}]{10.1145/3626772.3657664}
Joachim Vanneste, Manisha Verma, and Debasis Ganguly.
\newblock Detecting and explaining emotions in video advertisements.
\newblock In {\em Proceedings of the 47th International ACM SIGIR Conference on Research and Development in Information Retrieval}, SIGIR '24, page 2734–2738, New York, NY, USA, 2024. Association for Computing Machinery.

\bibitem[\protect\citeauthoryear{Zeiler and Fergus}{2013}]{zeiler2013visualizingunderstandingconvolutionalnetworks}
Matthew~D Zeiler and Rob Fergus.
\newblock Visualizing and understanding convolutional networks, 2013.

\end{thebibliography}
\clearpage
\appendix

\section{Training parameters for the VGG11-based optical flow model}
\label{app:training_VGG11}
 Key hyperparameters were configured as follows: a fixed learning rate of 0.001, a batch size of 16, a maximum amount of 50 epochs, the Adam optimizer, and cross-entropy loss were selected to ensure stability and efficiency during training. Early stopping was implemented with a patience of 7 epochs. The training concluded at epoch 24 when early stopping criteria were met.

 \section{Xception model architecture}
\label{app:IBA_Xception}
\begin{table}[H]
    \centering
    \renewcommand{\arraystretch}{1.2}
    \resizebox{\columnwidth}{!}{%
    \begin{tabular}{|p{2.5cm}|p{5cm}|}
    \hline
    \textbf{Layer} & \textbf{Type and Details} \\
    \hline\hline
    \textbf{conv1} & Conv2d, stride=(2,2), bias=False \\
    \hline
    \textbf{bn1} & BatchNorm2d, eps=1e-05, momentum=0.1 \\
    \hline
    \textbf{relu} & ReLU, inplace=True \\
    \hline\hline
    \textbf{conv2} & Conv2d, stride=(1,1), bias=False \\
    \hline
    \textbf{bn2} & BatchNorm2d, eps=1e-05, momentum=0.1 \\
    \hline
    \multicolumn{2}{|c|}{\textbf{Block Layers Repeated (block1 to block12)}} \\
    \hline\hline
    \textbf{conv3} & SeparableConv2d, Groups=1024, bias=False \\
    \hline
    \textbf{bn3} & BatchNorm2d, eps=1e-05, momentum=0.1 \\
    \hline\hline
    \textbf{conv4} & SeparableConv2d, Groups=1536, bias=False \\
    \hline
    \textbf{bn4} & BatchNorm2d, eps=1e-05, momentum=0.1 \\
    \hline
    \textbf{last\_linear} & Sequential, Fully Connected + Dropout \\
    \hline
    \end{tabular}}
    \caption{Xception model architecture (compact version).}
    \label{tab:Xception}
\end{table}

\begin{table}[H]
    \centering
    \renewcommand{\arraystretch}{1.2}
    \resizebox{\columnwidth}{!}{%
    \begin{tabular}{|p{2.5cm}|p{5cm}|}
    \hline
    \textbf{Layer} & \textbf{Type and Details} \\
    \hline\hline
    \textbf{0} & ReLU, inplace=True \\
    \hline
    \textbf{1} & SeparableConv2d, bias=False \\
    \hline
    \textbf{2} & BatchNorm2d, eps=1e-05, momentum=0.1 \\
    \hline
    \textbf{3} & ReLU, inplace=True \\
    \hline\hline
    \multicolumn{2}{|c|}{\textbf{Layers 1, 2, 3 can repeat here}} \\
    \hline\hline
    \textbf{4} & SeparableConv2d, bias=False \\
    \hline
    \textbf{5} & BatchNorm2d, eps=1e-05, momentum=0.1 \\
    \hline
    \textbf{6} & MaxPool2d, kernel\_size=3, stride=2, padding=1 \\
    \hline
    \end{tabular}}
    \caption{Typical architecture of the blocks with input/output channel amounts. Size values differ between different blocks.}
    \label{tab:block}
\end{table}

\section{VGG11-based model architecture}
\label{app:IBA_VGG11}

\renewcommand{\arraystretch}{1.0} 
\setlength{\tabcolsep}{2pt} 

\begin{table}[H]
    \centering
    \footnotesize 
    \begin{tabular}{|p{0.8cm}|p{1.8cm}|p{1.2cm}|p{1.2cm}|p{2.5cm}|}
        \hline
        \textbf{Layer} & \textbf{Type} & \textbf{Input} & \textbf{Output} & \textbf{Info} \\
        \hline
        0 & Conv2d & 3 & 64 &  \\
        1 & BatchNorm2d & 64 & 64 & eps=1e-5 \\
        2 & ReLU & - & - & inplace=True \\
        3 & MaxPool2d & - & - &  \\
        \hline
        4 & Conv2d & 64 & 128 &  \\
        5 & BatchNorm2d & 128 & 128 & eps=1e-5 \\
        6 & ReLU & - & - & inplace=True \\
        7 & MaxPool2d & - & - &  \\
        \hline
        8 & Conv2d & 128 & 256 &  \\
        9 & BatchNorm2d & 256 & 256 & eps=1e-5 \\
        10 & ReLU & - & - & inplace=True \\
        \hline
        11 & Conv2d & 256 & 256 &  \\
        12 & BatchNorm2d & 256 & 256 & eps=1e-5 \\
        13 & ReLU & - & - & inplace=True \\
        14 & MaxPool2d & - & - &  \\
        \hline
        15 & Conv2d & 256 & 512 &  \\
        16 & BatchNorm2d & 512 & 512 & eps=1e-5 \\
        17 & ReLU & - & - & inplace=True \\
        \hline
        18 & Conv2d & 512 & 512 &  \\
        19 & BatchNorm2d & 512 & 512 & eps=1e-5 \\
        20 & ReLU & - & - & inplace=True \\
        21 & MaxPool2d & - & - &  \\
        \hline
        22 & Conv2d & 512 & 512 &  \\
        23 & BatchNorm2d & 512 & 512 & eps=1e-5 \\
        24 & ReLU & - & - & inplace=True \\
        \hline
        25 & Conv2d & 512 & 512 &  \\
        26 & BatchNorm2d & 512 & 512 & eps=1e-5 \\
        27 & ReLU & - & - & inplace=True \\
        28 & MaxPool2d & - & - &  \\
        \hline
        last & Sequential & - & - & FC + Dropout \\
        linear &&&&\\
        \hline
    \end{tabular}
    \caption{VGG11 model architecture (compact format)}
    \label{tab:VGG11}
\end{table}

\newpage
\onecolumn
\section{Optical flow maps creation}
\label{app:optical_flow}
\begin{figure}[H]
\centering
\includegraphics[scale=0.65]{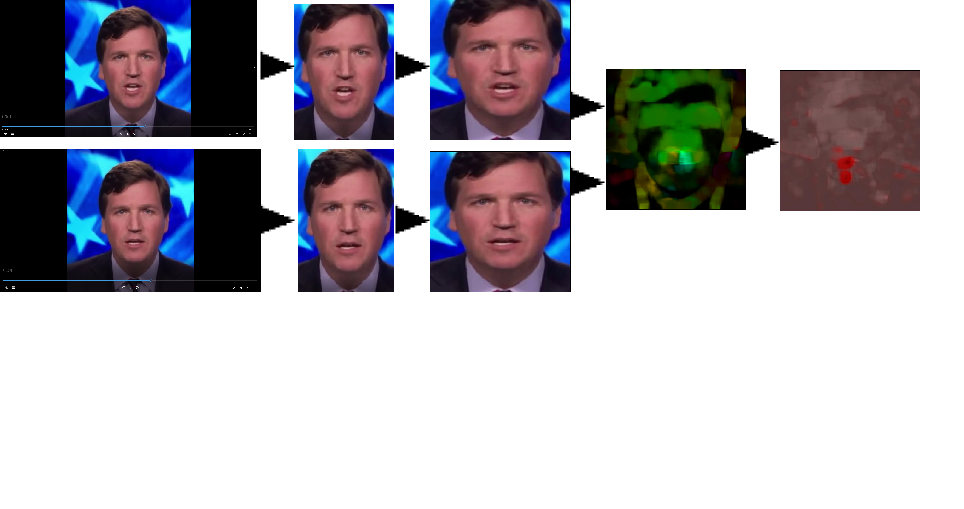}
\caption{Pre-processing pipeline for optical flow maps.}
\label{fig:opticalflow}
\end{figure}
\section{Examples of relevance maps generated with Xception and with bottleneck injection after different blocks}
\label{xception:choice}
\raggedbottom
\begin{figure}[H]
                \centering
                \includegraphics[scale=0.40]{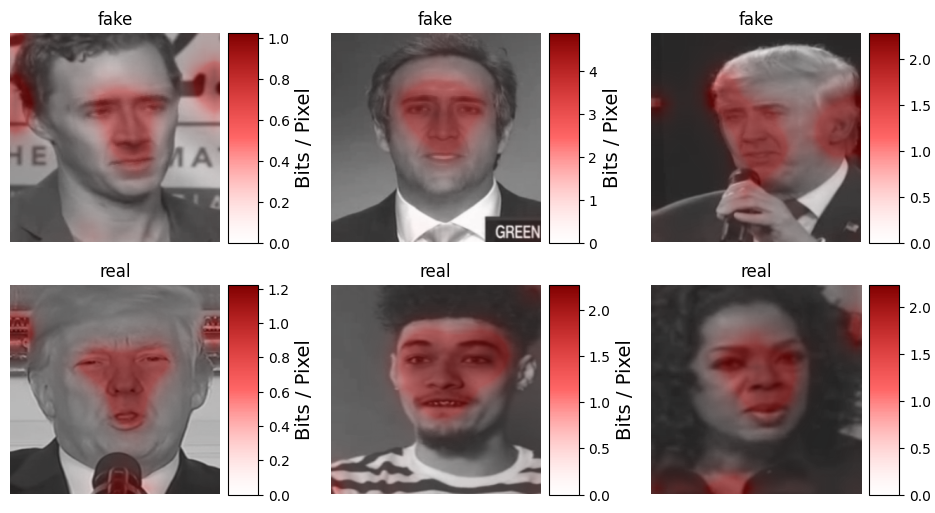}
                \caption{Relevance maps generated with Xception and with bottleneck injection after block 4.}
                \label{fig:block4}
            \end{figure}

            \begin{figure}[H]
                \centering
                \includegraphics[scale=0.40]{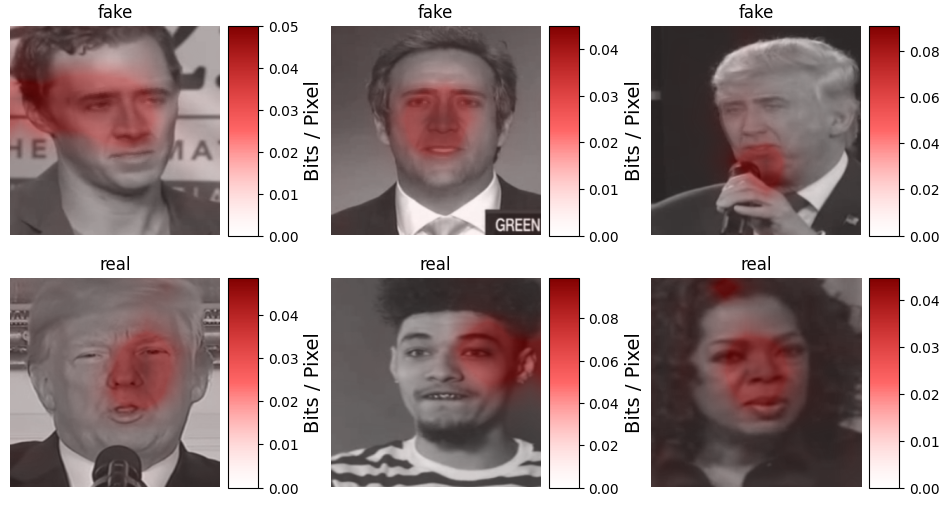}
                \caption{Relevance maps generated with Xception and with bottleneck injection after bn3.}
                \label{fig:bn3}
            \end{figure}

            \begin{figure}[H]
                \centering
                \includegraphics[scale=0.40]{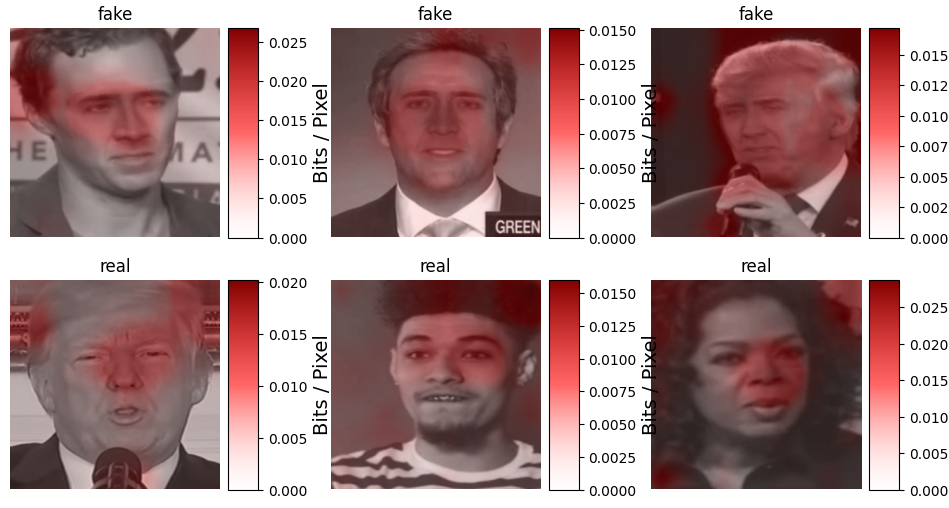}
                \caption{Relevance maps generated with Xception and with bottleneck injection after block 12.}
                \label{fig:block12}
            \end{figure}

            \begin{figure}[H]
                \centering
                \includegraphics[scale=0.40]{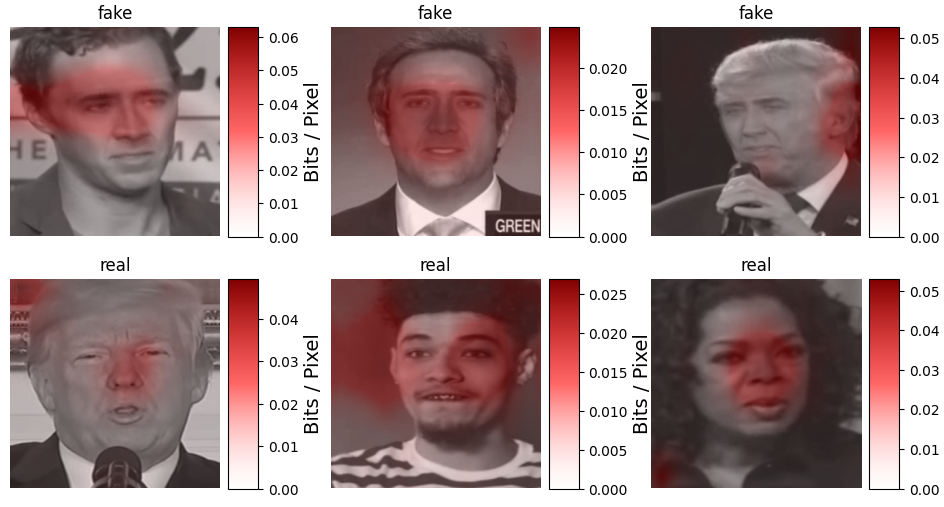}
                \caption{Relevance maps generated with Xception and with bottleneck injection after conv3.}
                \label{fig:conv3}
            \end{figure}

            \begin{figure}[H]
                \centering
                \includegraphics[scale=0.40]{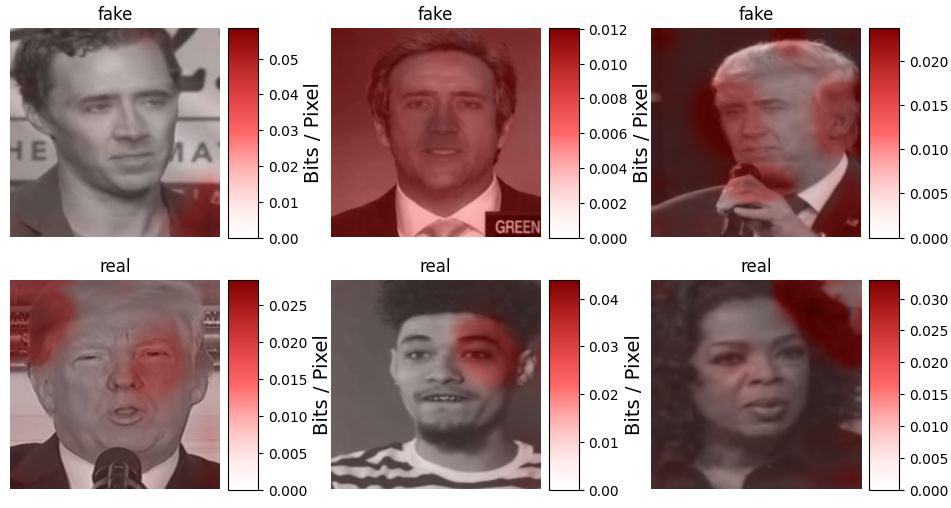}
                \caption{Relevance maps generated with Xception and with bottleneck injection after bn4.}
                \label{fig:bn4}
            \end{figure}
\section{Optical-flow maps generated with our VGG11-based model and with bottleneck injection after different layers.}
\label{vgg:choice}
\begin{figure}[H]
                \centering
                \includegraphics[scale=0.40]{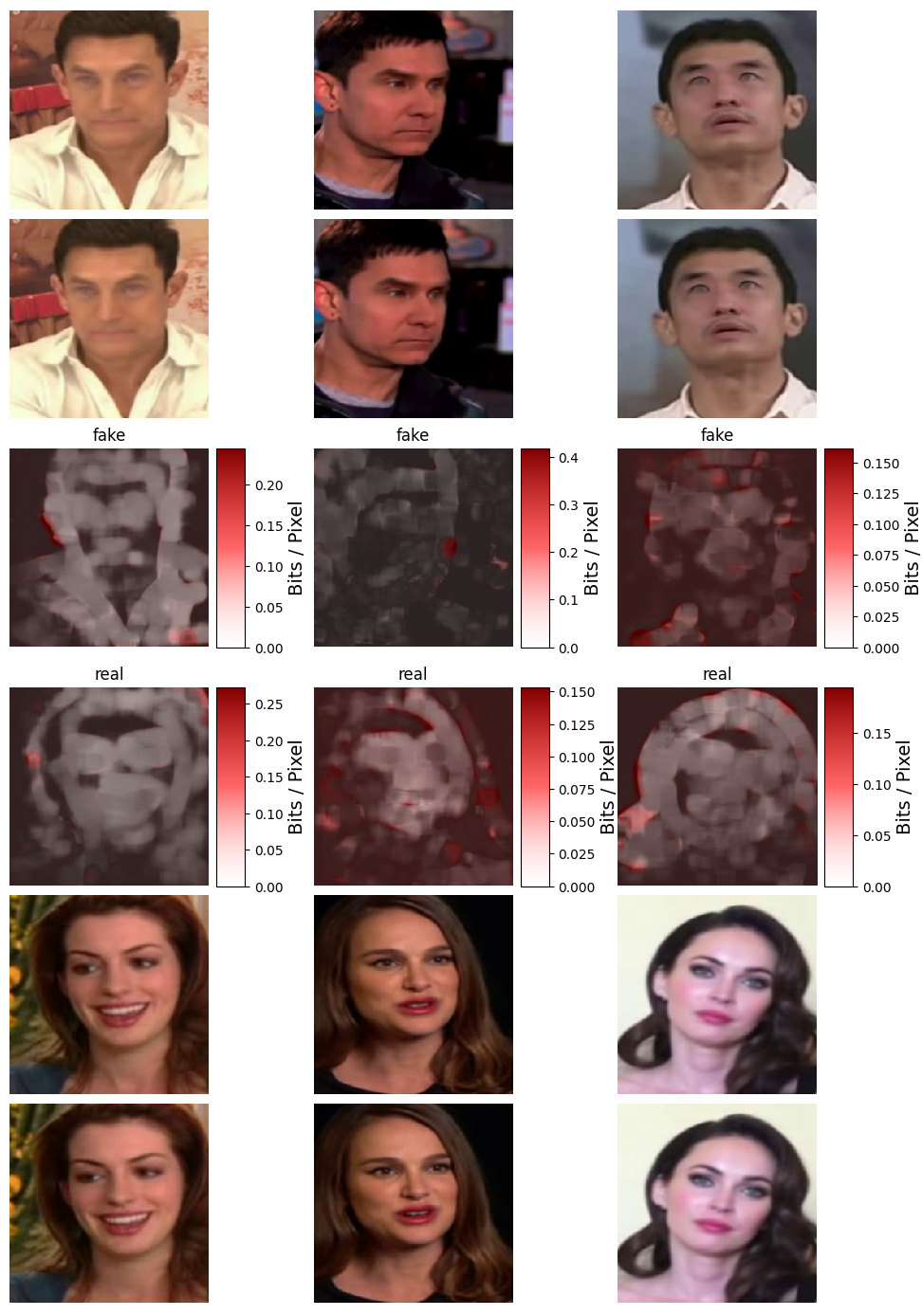}
                \caption{Relevance maps generated with VGG11 and bottleneck injection after layer 9, showing frame pairs used to create optical-flow maps}
                \label{fig:Batch Normalization2}
            \end{figure}

            \begin{figure}[H]
                \centering
                \includegraphics[scale=0.40]{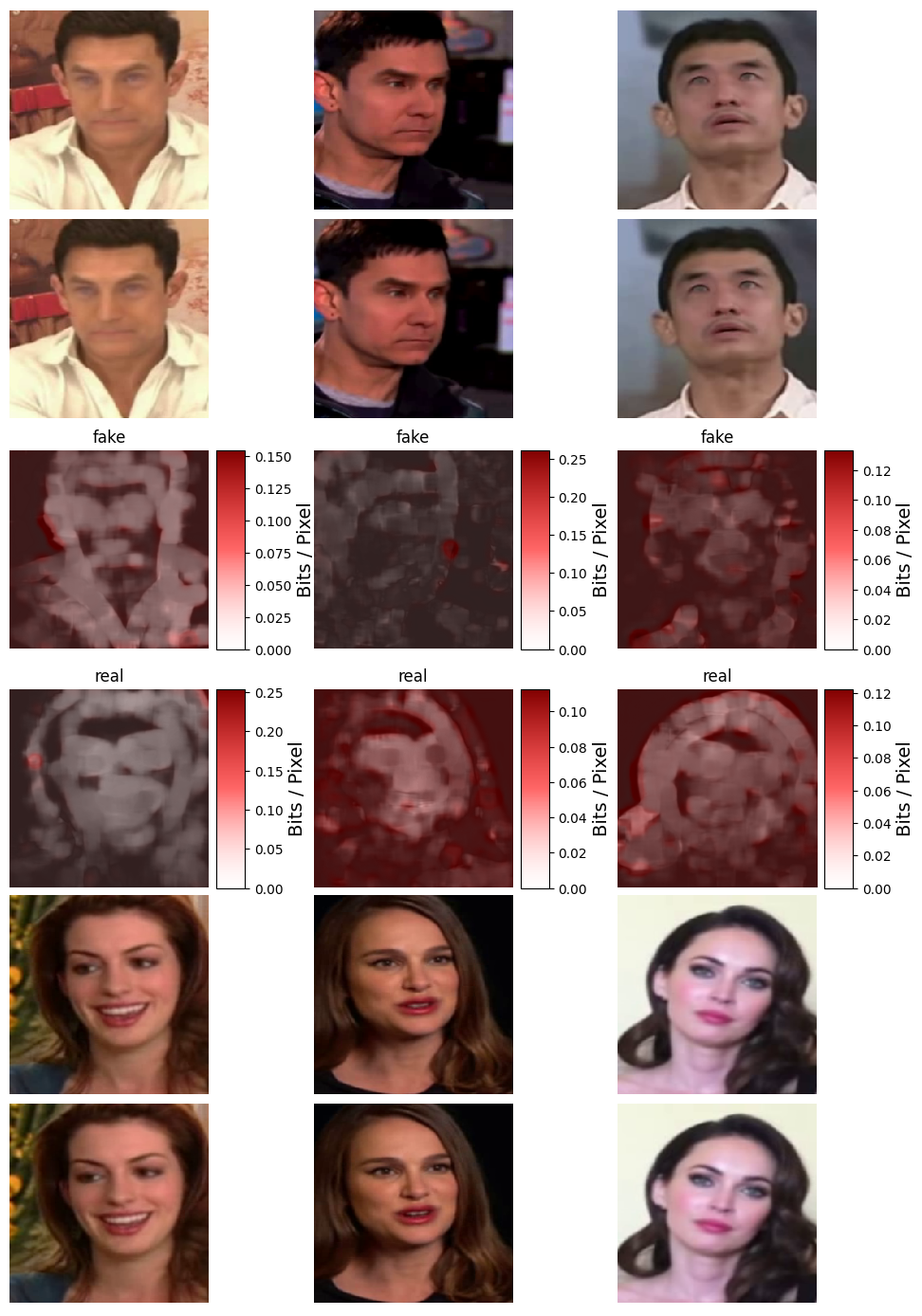}
                \caption{Relevance maps generated with VGG11 and bottleneck injection after layer 12, showing frame pairs used to create optical-flow maps}
                \label{fig:Batch Normalization3}
            \end{figure}

            \begin{figure}[H]
                \centering
                \includegraphics[scale=0.40]{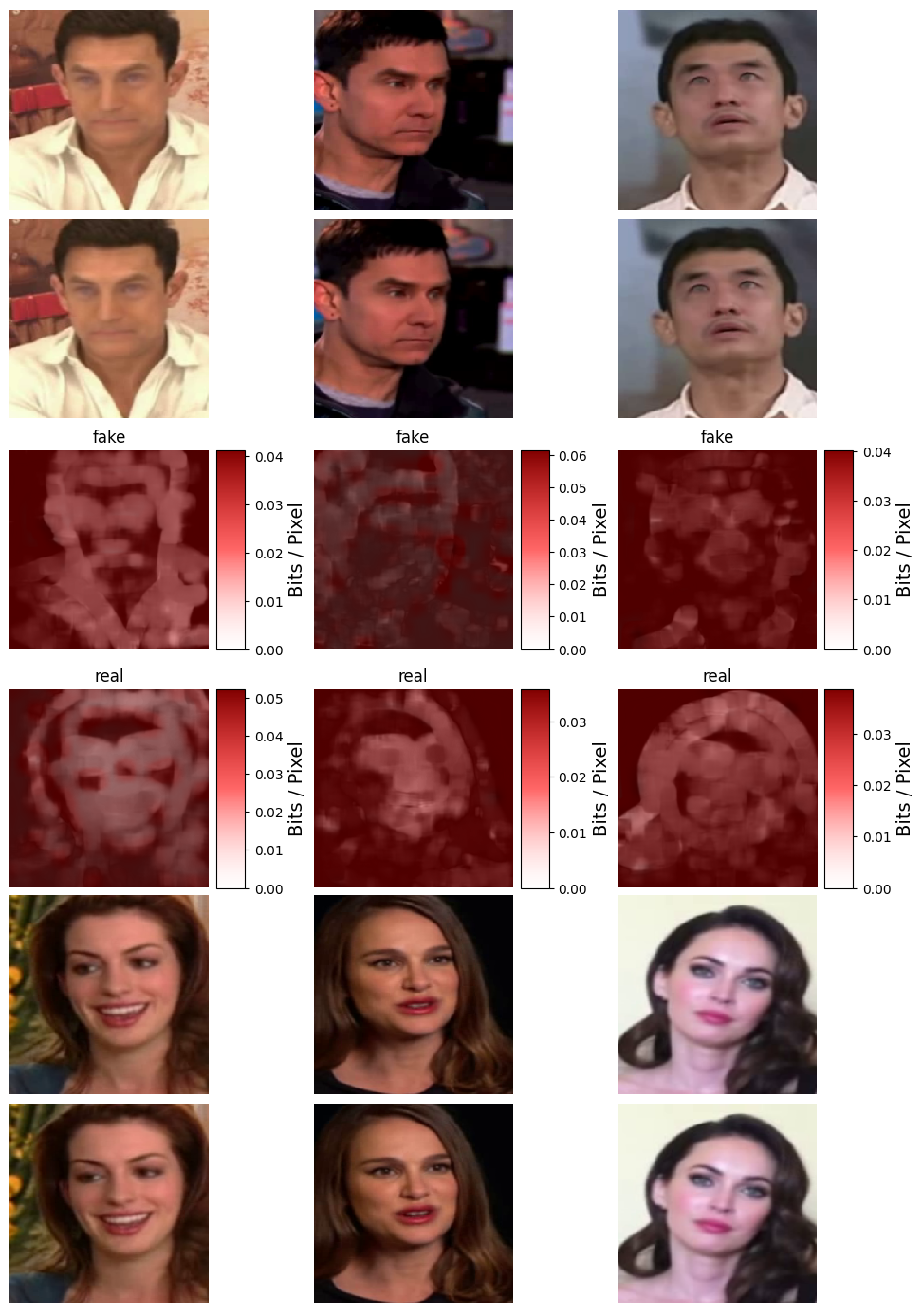}
                \caption{Relevance maps generated with VGG11 and bottleneck injection after layer 16, showing frame pairs used to create optical-flow maps}
                \label{fig:Batch Normalization4}
            \end{figure}

\end{document}